\documentclass[a4paper]{article}

\usepackage{INTERSPEECH2019}
\usepackage{multirow}
\usepackage{amsmath}
\usepackage{mathtools}

\DeclarePairedDelimiter\norm{\lVert}{\rVert}

\title{CNN-LSTM models for Multi-Speaker Source Separation using Bayesian Hyper Parameter Optimization
}
\name{Jeroen Zegers, Hugo Van hamme}
\address{
  KU Leuven, Dept. ESAT, Belgium}
\email{jeroen.zegers@esat.kuleuven.be, hugo.vanhamme@esat.kuleuven.be}

\begin{document}

\maketitle
\begin{abstract}
In recent years there have been many deep learning approaches towards the multi-speaker source separation problem. 
Most use Long Short-Term Memory - Recurrent Neural Networks (LSTM-RNN) or Convolutional Neural Networks (CNN) to model the sequential behavior of speech. In this paper we propose a novel network for source separation using an encoder-decoder CNN and LSTM in parallel. Hyper parameters have to be chosen for both parts of the network and they are potentially mutually dependent. Since hyper parameter grid search has a high computational burden, random search is often preferred. However, when sampling a new point in the hyper parameter space, it can potentially be very close to a previously evaluated point and thus give little additional information. Furthermore, random sampling is as likely to sample in a promising area as in an hyper space area dominated with poor performing models. Therefore, we use a Bayesian hyper parameter optimization technique and find that the parallel CNN-LSTM outperforms the LSTM-only and CNN-only model.


\end{abstract}
\noindent\textbf{Index Terms}: CNN-LSTM, Bayesian hyper parameter optimization, multi-speaker source separation
\section{Introduction}
Separating one speaker from another is a widely known speech processing problem, generally referred to as the cocktail party problem \cite{bregman1994auditory}. Multi-speaker source separation (MSSS) attempts to separate the signals of speakers when they speak simultaneously.
In recent years, many Deep Learning (DL) solutions have been proposed for the MSSS problem. Due to the sequential nature of speech, most approaches use (LSTM-)RNNs \cite{hershey2016deep,isik2016single,luo2018speaker,kolbaek2017multitalker,zegers2018multi}. However, there have also been solutions using CNNs \cite{kolbaek2017multitalker,li2018deep,DBLP:journals/corr/abs-1803-08629}. 

Most approaches use the Short Time Fourier Transform (STFT) as input to the network (although time-domain approaches also exist \cite{luo2018tasnet}), where the goal is to assign each time-frequency bin to a speaker (when making the sparsity assumption that in each bin, one speaker is dominant). The task can be related to the image segmentation problem where each pixel in the image has to be classified. The encoder-decoder CNN or SegNet can be used for this task \cite{Badrinarayanan2017}, which starts from high resolution low-level features, encoding them into low resolution high-level features in several steps. The task of the decoder is to bring the high-level information back to high resolution.

If SegNet is used for the MSSS task, it might be difficult for the network to extract the long term dependencies related to speaker identity that are important for the task \cite{zegers2018memory}. These long term effects span hundreds of time frames which would lead to either very wide convolutional kernels or extreme max-pooling. Therefore we add a LSTM network in parallel to the SegNet, to model the sequential behavior better.
To our knowledge CNN and LSTM have not been combined in parallel for the MSSS task before. Dilated convolutions are an alternative to deal with the long term dependencies \cite{DBLP:journals/corr/abs-1803-08629}.

A parallel CNN-LSTM has been proposed before for acoustic scene classification \cite{bae2016acoustic}. For that task the complete segment is labeled whereas for MSSS every time-frequency bin is labeled. This causes our network to be different from theirs: the CNN in \cite{bae2016acoustic} can be seen as the encoder part of SegNet, while our proposed network uses the encoder and the decoder part. Furthermore, the outputs of the LSTM and the CNN are combined in a different way in our proposal. Alternatively, the CNN and LSTM can be used in sequential order \cite{donahue2015long,sainath2015convolutional,zhao2019speech,XiaoC16,abs-1802-05630}. 

The CNN and the LSTM-RNN, as well as subsequent fully connected layers, have hyper parameters that have to be chosen. This can be an exhausting task since the hyper parameters are mutually dependent and evaluation of a hyper parameter set is expensive (and noisy). Hyper parameter grid search is computationally expensive and it has been shown that random search can achieve the same accuracy with less evaluations \cite{bergstra2012random}. However, random search does not estimate a distribution over the hyper parameter space. When a new hyper parameter set is chosen for evaluation, it is as likely to sample a point in a promising region of the hyper parameter space, as it is to sample in a region dominated by poor performing models. Furthermore, we would like to avoid a sample that is very close to a previously evaluated sample as it would provide little additional information. To model the distribution of the validation loss over the hyperspace we use Gaussian Processes (GP) \cite{williams2006gaussian}. When sampling, an acquisition function (e.g. probability of improvement) is used \cite{bergstra2011algorithms,snoek2012practical}.


The remainder of this paper is organized as follows. In section \ref{sec:MSSS} the MSSS task will be explained and in section \ref{sec:CNN-LSTM} the parallel CNN-LSTM model will be discussed. Experiments will be shown in section \ref{sec:exp} and a conclusion will be given in section \ref{sec:concl}.

\begin{figure*}[t]
  \centering
  \includegraphics[width=\linewidth]{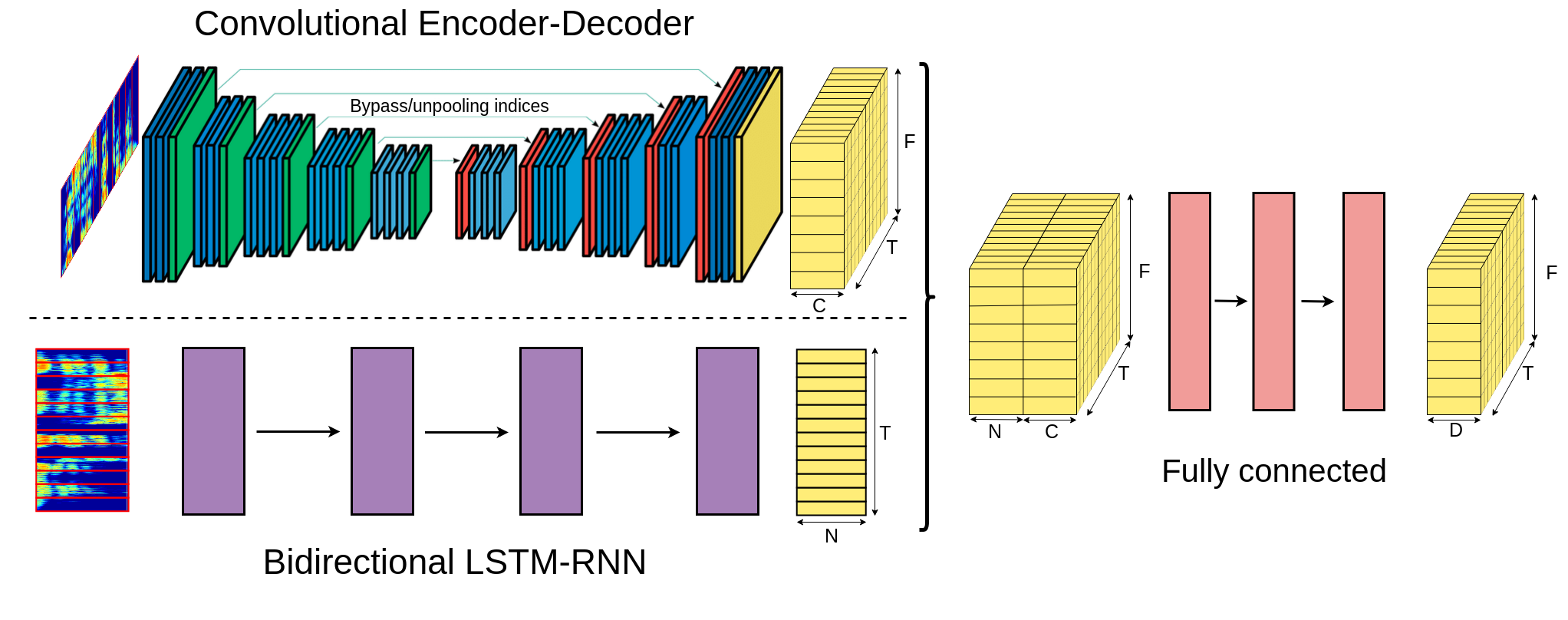}
  \caption{Parallel Convolutional Encoder-Decoder and Bidirectional LSTM-RNN using broadcast concatenation (based on \cite{Badrinarayanan2017}).}
  \label{fig:segnet}
\end{figure*}

\section{Multi Speaker Source Separation}
\label{sec:MSSS}
For the task of MSSS, signals $\hat{\mathbf{x}}_s[n]$ have to be estimated from a mixture $\mathbf{y}[n]=\sum_{s=1}^{S}\mathbf{x}_s[n]$ of $S$ speakers, where $\mathbf{x}_s[n]$ is the source signal for the $s^{th}$ speaker as recorded by the microphone. In the time-frequency domain, the same task can be expressed using the Short Time Fourier Transform (STFT) of the signals: Given a mixture $\mathbf{Y}(t,f)=\sum_{s=1}^{S}\mathbf{X}_s(t,f)$, estimate $\hat{\mathbf{X}}_s(t,f)$. Usually, a mask $\hat{\mathbf{M}}_s(t,f)$ is estimated for the $s^{th}$ speaker such that 

\begin{equation}
\label{eq:spec_est}
\hat{\mathbf{X}}_s(t,f)=\hat{\mathbf{M}}_s(t,f)\circ\mathbf{Y}(t,f)
\end{equation}
for every time frame $t=1, \ldots, T$ and every frequency $f=1, \ldots, F$ and with $\circ$ the Hadamard product \cite{Yu2017permutation}. The masks are constrained by $\hat{\mathbf{M}}_s(t,f)\ge 0$ and $\sum_{s=1}^{S}\hat{\mathbf{M}}_s(t,f)=1$ for every time-frequency bin $(t,f)$.

Deep Clustering (DC) is an often used method for MSSS \cite{hershey2016deep}. In DC a $D$-dimensional embedding vector $\mathbf{v}_{tf}$ is found for every $(t,f)$ via a mapping $\mathbf{v}_{tf}=f_{\theta}(\mathbf{Y})$. $f_{\theta}$ is based on a deep neural network and is chosen such that $\mathbf{v}_{tf}$ is normalized to unit length. The embedding vectors for every bin are stored as rows in a ($TF\times D$)-dimensional matrix $\mathbf{V}$. A ($TF\times S$)-dimensional target matrix $\mathbf{U}$ is defined such that $u_{tf,s}=1$ if target speaker $s$ has the most energy in the mixture for $(t,f)$ and $u_{tf,s}=0$ otherwise. A permutation independent loss function is then stated as
\begin{equation}
\label{eq:dc_loss}
\begin{split}
\mathcal{L}_{\theta} &= \norm{\mathbf{V}\mathbf{V}^T-\mathbf{U}\mathbf{U}^T}_F^2 \\
&= \sum_{t_1,f_1,t_2,f_2}(\langle \mathbf{v}_{t_1f_1},\mathbf{v}_{t_2f_2}\rangle-\langle \mathbf{u}_{t_1f_1},\mathbf{u}_{t_2f_2}\rangle)^2
\end{split}
\end{equation}
where $\norm{.}_F^2$ is the squared Frobenius norm. During inference, all embedding vectors are clustered into $S$ clusters $c$ using K-means. The masks are then constructed as follows
\begin{equation}
\hat{\mathbf{M}}_{s,tf}=
\begin{cases}
      1, & \text{if}\ \mathbf{v}_{tf} \in c_s \\
      0, & \text{otherwise}
    \end{cases}.
\end{equation}
Equation \ref{eq:spec_est} can then be used to estimate the original source signals via the inverse STFT and overlap-add \cite{rabiner1975theory}.

\section{CNN-LSTM Model}
\label{sec:CNN-LSTM}

\subsection{Encoder-Decoder CNN}
The encoder-decoder CNN or SegNet \cite{Badrinarayanan2017} is often used in image segmentation, where each pixel has to be labeled. The input has high resolution, low-level features and the encoder extracts high-level features, but in low resolution since max-pooling operations are used. The task of the decoder is to bring the high-level information to high resolution. However, it is difficult to convert a low resolution image to a high resolution image. Therefore resolution information from intermediate encoder layers is transferred over with the bypass mechanism or the unpooling operation (see figure \ref{fig:segnet}). The \emph{bypass} mechanism uses skip-connections from an encoding layer to the corresponding decoder layer. The \emph{unpooling} operation uses the positional information of the corresponding pooling operation in the encoder to reverses the pooling \cite{Badrinarayanan2017}.

\begin{table*}[th]
  \caption{Overview of hyper parameters and the allowed range. In column 4, 5 and 6 the values for the best performing CNN-LSTM model, LSTM-only model and the CNN-only model are given, respectively.}
  \label{tab:hyp_par}
  \centering
  \begin{tabular}{l | l | l | c | c | c }
    \toprule
    Network & Hyper parameter & range & CNN-LSTM & LSTM & CNN \\
    \midrule
    \multirow{ 16}{*}{CNN} & Nr. of encoder layers (= nr. of decoder layers) & 0-6 & 1 & 0 & 4            \\
     & Nr. of channels in the first encoder layer & 1-2000 & 489 & / & 74            \\
     & Increase factor in nr. of channels per layer ($C_l=C_1*fac^{l-1}$) & 0.5-3.0 & / & / & 1.48             \\
     & Nr. of channels in last decoder layer & 1-200 & 144 & / & 36            \\
     & Length of convolutive kernel in time dimension ($w_t$) & 1-15 & 2 & / & 14            \\
     & Decrease in length of convolutive kernel in time dimension & \multirow{ 2}{*}{0.83-3.33} & \multirow{ 2}{*}{/} & \multirow{ 2}{*}{/} & \multirow{ 2}{*}{0.79}             \\
     & after max pooling along the time dimension & & & & \\
     & Length of convolutive kernel in frequency dimension ($w_f$) & 1-15 & 12 & / & 8            \\
     & Decrease in length of convolutive kernel in frequency dimension & \multirow{ 2}{*}{0.83-3.33} &  \multirow{ 2}{*}{/} & \multirow{ 2}{*}{/} & \multirow{ 2}{*}{1.19}             \\
     &  after max pooling along the frequency dimension & & & & \\
     & Max pooling frequency along time dimension & \multirow{ 2}{*}{1-7} & \multirow{ 2}{*}{6} &  \multirow{ 2}{*}{/}& \multirow{ 2}{*}{2}             \\
     & (if 1, apply after every layer. If 2, apply after every 2 layers, etc.) & & & & \\
     & Max pooling frequency along frequency dimension & 1-7 &  2 & / & 3          \\
     & \multirow{ 3}{*}{Upsampling strategy} & \{bypass, &  \multirow{ 3}{*}{/} & \multirow{ 3}{*}{/} & \multirow{ 3}{*}{none}             \\
     &  & unpooling & &  &             \\
     &  & none\} & &  &             \\
    \midrule
    \multirow{ 4}{*}{LSTM} & Nr. of layers & 0-6 & 6 & 4 & 0           \\
     & Nr. of cells in the first layer & 1-2000 & 533 & 538 & /           \\
     & Increase factor in nr. of cells per layer & 0.5-3.0 & 0.73 & 0.79 & /             \\
     & Uni-directional or bi-directional & \{uni, bi\} & bi & bi & /             \\
    \midrule
    \multirow{ 5}{*}{FC}  & \multirow{ 2}{*}{Concatenate strategy} & \{broadcast, & \multirow{ 2}{*}{broadcast} & \multirow{ 2}{*}{/} & \multirow{ 2}{*}{broadcast}            \\
     & & flattening\} & &  &             \\
    & Nr. of layers (besides the output layer) & 0-3 & 2 & 1 & 1            \\
     & Nr. of units in the first layer & 1-1024 & 731 & 570 & 708            \\
     & Increase factor in nr. of units per layer & 0.3-2.0 & 0.44 & / & /            \\
    
    \bottomrule
  \end{tabular}
  
\end{table*}

\subsection{Bi-directional LSTM-RNN}
To cover long-term information, the length of the convolution kernel along the time dimension should be large or there should be many-max pooling operations. The former is unwanted due to the increase in trainable parameters and the latter decreases the time resolution of the encoded information. Therefore an LSTM-RNN is included, which is better suited to cope with long-term information. The LSTM can be uni-directional or bi-directional.

\subsection{Parallel CNN-LSTM}
A LSTM-RNN is added in parallel to the SegNet, meaning that it operates on the same input as the CNN and outputs of the LSTM and CNN are concatenated. The CNN output is denoted as a $T \times F \times C$ tensor $\mathbf{V}_{CNN}$, with $C$ the number of channels in the last decoder layer, while the LSTM output $\mathbf{V}_{LSTM}$ has shape $T \times N$, with $N$ the number of cells in the last LSTM layer. Two options to concatenate $\mathbf{V}_{CNN}$ and $\mathbf{V}_{LSTM}$ into a tensor $\mathbf{V}_{CNN\_LSTM}$ are considered. $\mathbf{V}_{LSTM}$ can be broadcast to the frequency dimension of $\mathbf{V}_{CNN}$ such that $\mathbf{V}_{CNN\_LSTM}$ has shape $T \times F \times (C+N)$. This is called the \emph{broadcast} option for further reference. Alternatively, the last two dimensions of $\mathbf{V}_{CNN}$ can be flattened so that $\mathbf{V}_{CNN\_LSTM}$ has shape $T \times (FC+N)$. This will be referred to as the \emph{flattening} option.

A Fully Connected (FC) network and an output layer are used to transform $\mathbf{V}_{CNN\_LSTM}$ into $\mathbf{V}$ with shape $T \times F \times D$, as requested by the Deep Clustering. If the broadcast option is used, weight are shared over the time and frequency dimension and every $(C+N)$-dimensional vector is transformed to a $D$-dimensional embedding and no further reshaping is needed. If the flattening option is chosen, weights are shared over the time dimension only and every $(FC+N)$-dimensional vector is transformed to a $FD$-dimensional vector and then reshaped to a $F \times D$ matrix.

\subsection{Hyper parameter optimization}
To allow for diverse architectures, many hyper parameters were chosen with a broad range of values. In table \ref{tab:hyp_par} an overview of all hyper parameters, as well as their range, is given. 



To find the optimal set of hyper parameters, we use Gaussian Processes (GP) combined with acquisition functions. The GP is used to model the distribution of the validation loss over the hyper space. For a detailed explanation of GP, see \cite{williams2006gaussian}. When looking for a new sample to evaluate, we would like to maximize the probability that the new sample will improve over the best validation loss so far. Acquisition functions make a trade-off when choosing where to sample. A region where few samples have yet been evaluated (high variance by the GP) is preferred as well as a region dominated by well performing samples (low predicted validation loss by the GP). More information on differences between acquisition functions can be found in \cite{snoek2012practical}. To find the maximum of the acquisition function, we used the L-BFGS algorithm \cite{byrd1995limited}.

\begin{figure*}
  \centering
  \includegraphics[width=1\linewidth]{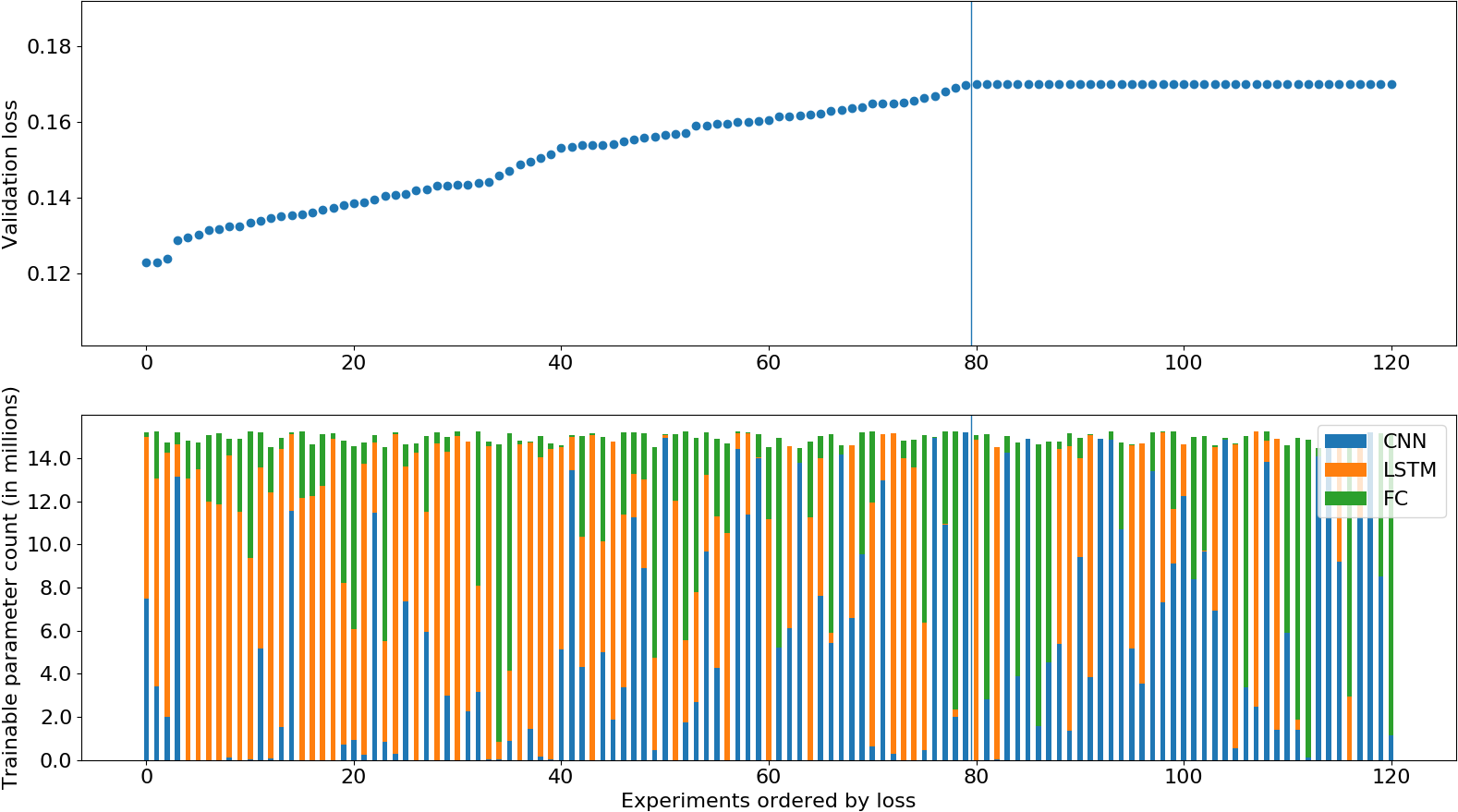}
  \caption{The top figure shows the experiments ordered by the validation loss. Experiments with validation loss above 0.17 are not ordered. The bottom figure shows the number of trainable parameters for the CNN, the LSTM and the FC.}
  \label{fig:all_losses}
\end{figure*}

\section{Experiments}
\label{sec:exp}
\subsection{Experimental setup}
Experiments were done on two speaker mixtures, artificially created by single speaker utterances from the Wall Street Journal 0 (WSJ0) corpus, sampled at 8kHz \cite{hershey2016deep}. The training and validation sets contained 20,000 and 5,000 mixtures, respectively and were taken from the \texttt{si\_tr\_s} set of WSJ0. The test set contained 3,000 mixtures using 16 held-out speakers of the \texttt{si\_dt\_05} and \texttt{si\_et\_05} set. The decimal log-magnitude of the STFT with a 32 ms window length and a hop size of 8 ms were used as features and were normalized with mean and variance, calculated over the whole training set.

All models have between 14.4 million and 15.2 million trainable parameters to make a fair comparison. The network parameters were trained with the Adam learning algorithm \cite{kingma2014adam} and early stopping on the validation set was used. Curriculum learning was used \cite{bengio2009curriculum}, i.e. the network was presented an easier task before tackling the main task. Here, the network was first trained on segments of 100 frames. Then 500-frame segments were used and finally the complete mixtures. Zero mean Gaussian noise with standard deviation $0.2$ was applied to the training features to avoid local optima. For DC the embedding dimension was chosen at $D=20$. Performance for MSSS was measured on the average signal-to-distortion ratio (SDR) improvements on the test set, using the \texttt{bss\_eval} toolbox \cite{vincent2006performance}. All networks were trained using TensorFlow \cite{abadi2016tensorflow} and hyper parameter optimization was done with the Scikit Optimize Python library \cite{Scikit}. The code for all the experiments can be found here: \\\texttt{https://github.com/JeroenZegers/Nabu-MSSS}.

\subsection{Results}

The optimally found hyper parameters for the CNN-LSTM, LSTM and CNN are shown in table \ref{tab:hyp_par}. Their validation loss and SDR score are shown in table \ref{tab:scores}. The CNN-LSTM outperforms both the LSTM and CNN in terms of SDR improvement. An additional LSTM was trained, called \textit{LSTM-der}, which was derived from the best performing CNN-LSTM model, but with the CNN part removed. The number of units in the LSTM model was increased to counter the reduction in trainable parameters. The CNN-LSTM also outperforms this model.

In the top part of figure \ref{fig:all_losses} all experiments are ordered by their validation loss, clipped at 0.17. This was done such that a model could be stopped early, with the purpose of saving computation time, when its validation loss was converging to a value significantly higher than 0.17. Consequently, all experiments with validation loss greater than or equal to 0.17 are not ordered in figure \ref{fig:all_losses}. The bottom part of figure \ref{fig:all_losses} shows the number of trainable parameters for all experiments as well as the partitioning into the CNN, LSTM and FC part of the network. The total amount of trainable parameters is fixed between 14.4 million and 15.2 million for fair comparison. 
It can be seen that the overall trend is that more trainable parameters in the LSTM part of the network is beneficial, but the best performing models use the CNN and LSTM in parallel. We do notice that generally the LSTM-only models outperform the CNN-only models.

\begin{table}[t]
  \caption{Results for the optimal found hyper parameters (see table \ref{tab:hyp_par}).}
  \label{tab:scores}
  \centering
  \begin{tabular}{c | c  c  c  c }
    \toprule
    Model & CNN-LSTM & LSTM & LSTM-der & CNN \\
    \midrule
    SDR & \textbf{9.82} & 9.62 & 9.59 & 7.99 \\
    Val. loss & \textbf{0.124} & 0.130 & 0.130 & 0.167 \\
    \bottomrule
  \end{tabular}
\end{table}

\begin{table}[th]
  \caption{Average results over the best 5 experiments for the different hyper parameter choices. For the hyper parameters \emph{'upsampling'} and \emph{'concatenate'}, only experiments where at least 20\% of the trainable parameters were part of the CNN were considered. For hyper parameter \emph{'LSTM direction'}, only experiments where at least 20\% of the trainable parameters were part of the LSTM were considered }
  \label{tab:scores_hyp}
  \centering
  \begin{tabular}{c c | c   c }
    \toprule
    \multicolumn{2}{c|}{Hyper parameter} & SDR & Validation loss \\
    \midrule
    \multirow{3}{*}{Upsampling} & bypass & \textbf{9.12} & \textbf{0.132} \\
         & unpool & 8.51 & 0.144  \\
         & none & 8.12 & 0.155\\
    \midrule
    \multirow{2}{*}{Concatenate} & broadcast& \textbf{9.26}  & \textbf{0.129} \\
         & flattening & 8.17 & 0.150 \\
    \midrule
    \multirow{2}{*}{LSTM direction} & uni & 8.09 & 0.153 \\
         & bi & \textbf{9.63}  & \textbf{0.126} \\
    \bottomrule
  \end{tabular}
\end{table}

In table \ref{tab:scores_hyp} we give a closer analysis for some categorical hyper parameters. For each category the five best performing models were selected and the scores were averaged. For the \emph{upsampling} and \emph{concatenate} hyper parameters, only models where at least 20\% of the trainable parameters were part of the CNN were considered. For \emph{LSTM direction}, only models where at least 20\% of the trainable parameters were part of the LSTM were considered. We notice that \emph{bypass} is the preferred upsampling strategy over \emph{unpooling} or using no upsampling. The \emph{broadcast} concatenate strategy is preferred of the \emph{flattening} strategy. Finally, we see that \emph{bi-directional} LSTMs are preferred.

\section{Conclusions}
\label{sec:concl}
By using Bayesian hyper parameter optimization, we found that parallel CNN-LSTM models outperform LSTMs and CNNs. In general, models with more trainable parameters in the LSTM part of the model are preferred. In further work, the hyper parameter optimization technique can be extended to different architectures like sequential CNN-LSTM or parallel CNN-LSTM with intermediate connections.

\section{Acknowledgements}
This work was funded by the SB PhD grant of the Research Foundation Flanders
(FWO) with project number 1S66217N.

\newpage
\bibliographystyle{IEEEtran}

\bibliography{mybib}

\begin{thebibliography}{10}
\providecommand{\url}[1]{#1}
\csname url@samestyle\endcsname
\providecommand{\newblock}{\relax}
\providecommand{\bibinfo}[2]{#2}
\providecommand{\BIBentrySTDinterwordspacing}{\spaceskip=0pt\relax}
\providecommand{\BIBentryALTinterwordstretchfactor}{4}
\providecommand{\BIBentryALTinterwordspacing}{\spaceskip=\fontdimen2\font plus
\BIBentryALTinterwordstretchfactor\fontdimen3\font minus
  \fontdimen4\font\relax}
\providecommand{\BIBforeignlanguage}[2]{{%
\expandafter\ifx\csname l@#1\endcsname\relax
\typeout{** WARNING: IEEEtran.bst: No hyphenation pattern has been}%
\typeout{** loaded for the language `#1'. Using the pattern for}%
\typeout{** the default language instead.}%
\else
\language=\csname l@#1\endcsname
\fi
#2}}
\providecommand{\BIBdecl}{\relax}
\BIBdecl

\bibitem{bregman1994auditory}
A.~S. Bregman, \emph{Auditory scene analysis: The perceptual organization of
  sound}.\hskip 1em plus 0.5em minus 0.4em\relax MIT press, 1994.

\bibitem{hershey2016deep}
J.~R. Hershey, Z.~Chen, J.~Le~Roux, and S.~Watanabe, ``Deep clustering:
  Discriminative embeddings for segmentation and separation,'' in \emph{2016
  IEEE International Conference on Acoustics, Speech and Signal Processing
  (ICASSP)}.\hskip 1em plus 0.5em minus 0.4em\relax IEEE, 2016, pp. 31--35.

\bibitem{isik2016single}
Y.~Isik, J.~L. Roux, Z.~Chen, S.~Watanabe, and J.~R. Hershey, ``Single-channel
  multi-speaker separation using deep clustering,'' \emph{arXiv preprint
  arXiv:1607.02173}, 2016.

\bibitem{luo2018speaker}
Y.~Luo, Z.~Chen, and N.~Mesgarani, ``Speaker-independent speech separation with
  deep attractor network,'' \emph{IEEE/ACM Transactions on Audio, Speech, and
  Language Processing}, vol.~26, no.~4, pp. 787--796, 2018.

\bibitem{kolbaek2017multitalker}
M.~Kolb{\ae}k, D.~Yu, Z.-H. Tan, J.~Jensen, M.~Kolbaek, D.~Yu, Z.-H. Tan, and
  J.~Jensen, ``Multitalker speech separation with utterance-level permutation
  invariant training of deep recurrent neural networks,'' \emph{IEEE/ACM
  Transactions on Audio, Speech and Language Processing (TASLP)}, vol.~25,
  no.~10, pp. 1901--1913, 2017.

\bibitem{zegers2018multi}
J.~Zegers and H.~Van~hamme, ``Multi-scenario deep learning for multi-speaker
  source separation,'' in \emph{2018 IEEE International Conference on
  Acoustics, Speech and Signal Processing (ICASSP)}.\hskip 1em plus 0.5em minus
  0.4em\relax IEEE, 2018, pp. 5379--5383.

\bibitem{li2018deep}
L.~Li and H.~Kameoka, ``Deep clustering with gated convolutional networks,'' in
  \emph{2018 IEEE International Conference on Acoustics, Speech and Signal
  Processing (ICASSP)}.\hskip 1em plus 0.5em minus 0.4em\relax IEEE, 2018, pp.
  16--20.

\bibitem{DBLP:journals/corr/abs-1803-08629}
\BIBentryALTinterwordspacing
S.~Mobin, B.~Cheung, and B.~A. Olshausen, ``Convolutional vs. recurrent neural
  networks for audio source separation,'' \emph{CoRR}, vol. abs/1803.08629,
  2018. [Online]. Available: \url{http://arxiv.org/abs/1803.08629}
\BIBentrySTDinterwordspacing

\bibitem{luo2018tasnet}
Y.~Luo and N.~Mesgarani, ``Tasnet: time-domain audio separation network for
  real-time, single-channel speech separation,'' in \emph{2018 IEEE
  International Conference on Acoustics, Speech and Signal Processing
  (ICASSP)}.\hskip 1em plus 0.5em minus 0.4em\relax IEEE, 2018, pp. 696--700.

\bibitem{Badrinarayanan2017}
V.~Badrinarayanan, A.~Kendall, and R.~Cipolla, ``Segnet: A deep convolutional
  encoder-decoder architecture for image segmentation,'' \emph{IEEE
  transactions on pattern analysis and machine intelligence}, vol.~39, pp.
  2481--2495, 2017.

\bibitem{zegers2018memory}
J.~Zegers and H.~Van~hamme, ``Memory time span in lstms for multi-speaker
  source separation,'' in \emph{Proceedings of the Annual Conference of the
  International Speech Communication Association, INTERSPEECH}, vol.
  2018.\hskip 1em plus 0.5em minus 0.4em\relax ISCA, 2018, pp. 1477--1481.

\bibitem{bae2016acoustic}
S.~H. Bae, I.~Choi, and N.~S. Kim, ``Acoustic scene classification using
  parallel combination of lstm and cnn,'' in \emph{Proceedings of the Detection
  and Classification of Acoustic Scenes and Events 2016 Workshop (DCASE2016)},
  2016, pp. 11--15.

\bibitem{donahue2015long}
J.~Donahue, L.~Anne~Hendricks, S.~Guadarrama, M.~Rohrbach, S.~Venugopalan,
  K.~Saenko, and T.~Darrell, ``Long-term recurrent convolutional networks for
  visual recognition and description,'' in \emph{Proceedings of the IEEE
  conference on computer vision and pattern recognition}, 2015, pp. 2625--2634.

\bibitem{sainath2015convolutional}
T.~N. Sainath, O.~Vinyals, A.~Senior, and H.~Sak, ``Convolutional, long
  short-term memory, fully connected deep neural networks,'' in
  \emph{Acoustics, Speech and Signal Processing (ICASSP), 2015 IEEE
  International Conference on}.\hskip 1em plus 0.5em minus 0.4em\relax IEEE,
  2015, pp. 4580--4584.

\bibitem{zhao2019speech}
J.~Zhao, X.~Mao, and L.~Chen, ``Speech emotion recognition using deep 1d \& 2d
  cnn lstm networks,'' \emph{Biomedical Signal Processing and Control},
  vol.~47, pp. 312--323, 2019.

\bibitem{XiaoC16}
\BIBentryALTinterwordspacing
Y.~Xiao and K.~Cho, ``Efficient character-level document classification by
  combining convolution and recurrent layers,'' \emph{CoRR}, vol.
  abs/1602.00367, 2016. [Online]. Available:
  \url{http://arxiv.org/abs/1602.00367}
\BIBentrySTDinterwordspacing

\bibitem{abs-1802-05630}
\BIBentryALTinterwordspacing
C.~Etienne, G.~Fidanza, A.~Petrovskii, L.~Devillers, and B.~Schmauch, ``Speech
  emotion recognition with data augmentation and layer-wise learning rate
  adjustment,'' \emph{CoRR}, vol. abs/1802.05630, 2018. [Online]. Available:
  \url{http://arxiv.org/abs/1802.05630}
\BIBentrySTDinterwordspacing

\bibitem{bergstra2012random}
J.~Bergstra and Y.~Bengio, ``Random search for hyper-parameter optimization,''
  \emph{Journal of Machine Learning Research}, vol.~13, no. Feb, pp. 281--305,
  2012.

\bibitem{williams2006gaussian}
C.~K. Williams and C.~E. Rasmussen, \emph{Gaussian processes for machine
  learning}.\hskip 1em plus 0.5em minus 0.4em\relax MIT Press Cambridge, MA,
  2006, vol.~2, no.~3.

\bibitem{bergstra2011algorithms}
J.~S. Bergstra, R.~Bardenet, Y.~Bengio, and B.~K{\'e}gl, ``Algorithms for
  hyper-parameter optimization,'' in \emph{Advances in neural information
  processing systems}, 2011, pp. 2546--2554.

\bibitem{snoek2012practical}
J.~Snoek, H.~Larochelle, and R.~P. Adams, ``Practical bayesian optimization of
  machine learning algorithms,'' in \emph{Advances in neural information
  processing systems}, 2012, pp. 2951--2959.

\bibitem{Yu2017permutation}
M.~Kolb{\ae}k, D.~Yu, Z.~Tan, and J.~Jensen, ``Multitalker speech separation
  with utterance-level permutation invariant training of deep recurrent neural
  networks,'' pp. 1901--1913, 2017.

\bibitem{rabiner1975theory}
L.~R. Rabiner and B.~Gold, ``Theory and application of digital signal
  processing,'' \emph{Englewood Cliffs, NJ, Prentice-Hall, Inc., 1975. 777 p.},
  1975.

\bibitem{byrd1995limited}
R.~H. Byrd, P.~Lu, J.~Nocedal, and C.~Zhu, ``A limited memory algorithm for
  bound constrained optimization,'' \emph{SIAM Journal on Scientific
  Computing}, vol.~16, no.~5, pp. 1190--1208, 1995.

\bibitem{kingma2014adam}
D.~Kingma and J.~Ba, ``Adam: A method for stochastic optimization,''
  \emph{arXiv preprint arXiv:1412.6980}, 2014.

\bibitem{bengio2009curriculum}
Y.~Bengio, J.~Louradour, R.~Collobert, and J.~Weston, ``Curriculum learning,''
  in \emph{Proceedings of the 26th annual international conference on machine
  learning}.\hskip 1em plus 0.5em minus 0.4em\relax ACM, 2009, pp. 41--48.

\bibitem{vincent2006performance}
E.~Vincent, R.~Gribonval, and C.~F{\'e}votte, ``Performance measurement in
  blind audio source separation,'' \emph{IEEE transactions on audio, speech,
  and language processing}, vol.~14, no.~4, pp. 1462--1469, 2006.

\bibitem{abadi2016tensorflow}
M.~Abadi, A.~Agarwal, P.~Barham, E.~Brevdo, Z.~Chen, C.~Citro, G.~S. Corrado,
  A.~Davis, J.~Dean, M.~Devin \emph{et~al.}, ``Tensorflow: Large-scale machine
  learning on heterogeneous distributed systems,'' \emph{arXiv preprint
  arXiv:1603.04467}, 2016.

\bibitem{Scikit}
``Scikit optimize,'' \url{https://scikit-optimize.github.io/}, accessed:
  2019-04-04.

\end{thebibliography}



\end{document}